\documentclass[letterpaper, 10 pt, conference]{ieeeconf}  

\IEEEoverridecommandlockouts                              

\usepackage{graphics} 
\usepackage{epsfig} 
\usepackage{mathrsfs}
\usepackage{times} 
\usepackage{amsmath} 
\usepackage{amssymb}  
\usepackage{graphicx}
\usepackage{tabularx,array,ragged2e,booktabs}
\usepackage{color}
\usepackage{lipsum}
\usepackage{subcaption}
\usepackage{booktabs}
\usepackage{xcolor}
\usepackage{pifont}
\newcommand{\cmark}{\ding{51}} 
\newcommand{\xmark}{\ding{55}} 
\newcolumntype{L}{>{$}l<{$}}
\usepackage{placeins}
\usepackage{multirow}

\usepackage[style=ieee,dashed=false]{biblatex}

\title{\LARGE \bf
\textsc{RoVerFly}: Robust and Versatile Implicit Hybrid \\ 
Control of Quadrotor-Payload Systems
}

\author{Mintae Kim\textsuperscript{†}, Jiaze Cai, and Koushil Sreenath
\thanks{\textsuperscript{†} Corresponding author (e-mail: \texttt{mintae.kim@berkeley.edu})}
\thanks{The authors are with \textit{Hybrid Robotics Lab}, University of California, Berkeley, CA 94720, United States.}
}

\begin{document}

\maketitle
\thispagestyle{empty}
\pagestyle{empty}


\begin{abstract}
Designing robust controllers for precise trajectory tracking with quadrotors is challenging due to nonlinear dynamics and underactuation, and becomes harder with flexible cable–suspended payloads that add degrees of freedom and hybrid dynamics. Classical model-based methods offer stability guarantees but require extensive tuning and often fail to adapt when the configuration changes—when a payload is added or removed, or when its mass or cable length varies. We present \textsc{RoVerFly}, a unified learning-based control framework where a single reinforcement learning (RL) policy functions as an implicit hybrid controller, managing complex dynamics without explicit mode detection or controller switching. Trained with task and domain randomization, the controller is resilient to disturbances and varying dynamics. It achieves strong zero-shot generalization across payload settings—including no payload as well as varying mass and cable length—without re-tuning, while retaining the interpretability and structure of a feedback tracking controller. Code and supplementary materials are available at \url{https://github.com/mintaeshkim/roverfly}.
\end{abstract}


\section{Introduction}

Quadrotors are widely used for aerial navigation, and numerous planning and control strategies have been developed for agile, precise maneuvering \cite{lee2010geometric, kaufmann2023champion}. For payload transport, rigid (active) attachment increases manipulation capability but reduces agility due to higher inertia \cite{thomas2013avian}, whereas cable suspension preserves maneuverability and enables multi-UAV cooperation at the cost of additional underactuation, nonlinear dynamics, and hybrid dynamics from cable slack/taut transitions \cite{sreenath2013geometric, lee2013geometric}.

Prior work on quadrotor–payload systems falls into three directions: (1) generating and tracking payload trajectories that suppress swing \cite{tang2015mixed, tang2020multi, foehn2017fast}; (2) controlling systems with elastic/flexible cables and pulley mechanisms \cite{kotaru2017elastic, kotaru2018flexible, zeng2019pulley}; and (3) cooperative manipulation with multiple quadrotors \cite{tang2020multi, kotaru2018flexible}. Most approaches use model-based geometric control leveraging differential flatness \cite{sreenath2013geometric, zeng2020differential}. While these methods offer theoretical guarantees, they require accurate models and extensive tuning, and often fail to generalize to dynamic or uncertain settings. A key limitation is the need for separate controller designs for different configurations (with vs.\ without payload) \cite{lee2010geometric, lee2018geometric}, which necessitates explicit switching in practice and can introduce instability. Moreover, reliance on flatness requires precomputed trajectories and precludes tracking dynamically infeasible references.

Model-free RL has emerged as a promising paradigm for UAV control, learning feedback policies directly from interaction without explicit dynamics modeling \cite{hwangbo2017control, kaufmann2023champion, huang2023datt, cai2024learning}. This is particularly compelling for UAVs, whose high-rate, nonlinear, underactuated, and aerodynamically coupled dynamics are difficult to model and tune accurately \cite{bauersfeld2024robotics, gupta2025estimation}. RL controllers have shown robust trajectory tracking \cite{kaufmann2020deep}, cross-platform adaptation \cite{zhang2023learning}, disturbance rejection \cite{zhang2024proxfly}, and tracking of dynamically infeasible trajectories \cite{huang2023datt}. However, existing methods adequately handle rapidly varying payload conditions such as changes in mass or cable length, nor do they offer a unified solution to the switching problem inherent in these hybrid systems.

Table~\ref{tab:comparison} summarizes the landscape. Racing-oriented learning controllers excel on fixed courses with specialized perception/estimation stacks but are not intended for arbitrary references or payload handling~\cite{kaufmann2023champion}. General-purpose RL benchmarking suites enable reproducible studies yet encourage task-specific rewards and typically train a separate policy per task rather than a single versatile controller~\cite{xu2024omnidrones}. Hybrid learning–plus–adaptive control can follow references and reject wind but relies on a nominal model and an additional adaptive layer, with demonstrations mainly on simple planar motions~\cite{huang2023datt}. Methods that adapt via latent extrinsics capture cross-platform variation near hover but require an adaptation phase and are evaluated chiefly under small disturbances~\cite{zhang2023learning}. Taken together, these limitations point to the need for a controller that tracks arbitrary references while accommodating unknown payload presence, cable length, and mass—without per-task policies or switching.

We introduce \textsc{RoVerFly}, a learning-based tracking controller whose core is a \emph{single} policy functioning as an \emph{implicit hybrid controller}. This data-driven approach directly addresses the key limitation of classical methods by learning to manage the varied, hybrid dynamics of the payload system—from no payload to a flexible suspended cable—without explicit mode detection or switching. By leveraging task and domain randomization, the policy becomes robust to underactuation, nonlinearities, and disturbances. Drawing on POMDP insights, we use I/O history as a belief-state proxy, enabling the policy to act as an implicit observer to infer latent system properties and make robust decisions under partial observability. The result is a single RL-learned feedback law that unifies hovering, takeoff, landing, and trajectory tracking while avoiding task-specific reward engineering or environment-specific tuning.

\noindent\textbf{Contributions.}
\textit{(i) Unified, implicit hybrid control framework:} A single policy functioning as an implicit hybrid controller for robust arbitrary trajectory tracking in both quadrotor-only and flexible cable-suspended payload systems, generalizing across payload mass and cable length without switching or fine-tuning.  
\textit{(ii) Data-driven control for flexible cable payloads:} A versatile learning-based controller targeting the complex underactuated dynamics of flexible cable-suspended payloads, with strong zero-shot performance achieved via task and domain randomization.  
\textit{(iii) Belief-based control using I/O history:} A justification for incorporating I/O history grounded in POMDPs, showing its role in enabling the policy to implicitly estimate the system's hybrid mode and state for robust control under partial observability.


\begin{table}[t]
    \centering
    \setlength{\tabcolsep}{4pt}
    \renewcommand{\arraystretch}{1.1}
    \caption{Comparison of existing learning-based quadrotor control approaches.}
    \label{tab:comparison}
    \resizebox{\linewidth}{!}{
    \begin{tabular}{@{}lccccc@{}}
    \toprule
    \textbf{Criteria} & \textbf{\cite{kaufmann2023champion}} & \textbf{\cite{xu2024omnidrones}} & \textbf{\cite{huang2023datt}} & \textbf{\cite{zhang2023learning}} & \textsc{RoVerFly} \\
    \midrule
    Arbitrary trajectory tracking        & \xmark & \xmark & \cmark & \xmark                      & \cmark \\
    Robustness under disturbances        & \xmark & \xmark & \cmark & $\triangle$ Small impulse   & \cmark \\
    Versatility to varying tasks         & --     & \xmark & --     & $\triangle$ Hovering only   & \cmark \\
    Adaptation to varying environments   & \xmark & \xmark & \xmark & \cmark                      & \cmark \\
    Agnostic to system dynamics          & \cmark & \cmark & \xmark & \cmark                      & \cmark \\
    \bottomrule
    \end{tabular}}
\end{table}


\section{Related Work}

\subsection{Model-Based Control and Trajectory Planning for Quadrotor-Payload Systems}

Model-based approaches—particularly geometric control—provide formulations with almost-global stability guarantees for quadrotor–payload systems \cite{sreenath2013geometric, tang2015mixed, tang2020multi, zeng2020differential}. Via differential flatness, they plan trajectories by expressing states and inputs as flat outputs, ensuring feasibility. Geometric controllers on this principle track both quadrotor and payload trajectories \cite{sreenath2013geometric, lee2010geometric}. Recent work extends these methods to elastic or flexible cables \cite{kotaru2017elastic, kotaru2018flexible}, introducing higher-order dynamics and added underactuation. However, they require accurate models, frequent retuning, and are sensitive to unmodeled effects (e.g., actuator saturation, disturbances), limiting robustness in practice.

Trajectory-generation methods likewise plan dynamically feasible payload motions. Optimization-based approaches such as mixed-integer quadratic programming (MIQP) \cite{tang2015mixed} and mathematical programs with complementarity constraints (MPCC) \cite{foehn2017fast} handle hybrid dynamics and cable-tension regimes for swing suppression and obstacle avoidance. MIQP is computationally heavy, while MPCC improves efficiency but remains model- and tuning-sensitive; as a result, these methods are typically used offline or in settings with limited adaptability.


\subsection{Learning-based Control of Quadrotors}

Model-free RL has become a practical tool for quadrotor flight, enabling high-performance trajectory tracking and agile maneuvers without explicit system identification. The learned policy acts as a nonlinear feedback controller adapting to complex dynamics from data. RL controllers show disturbance rejection when paired with adaptive elements \cite{huang2023datt}, improved sample efficiency and sim-to-real transfer with recent strategies \cite{eschmann2024learning}, and race-level performance where simulation-trained controllers surpassed human pilots \cite{kaufmann2023champion}.

Despite progress, payload trajectory tracking—especially with flexible, cable-suspended payloads—remains underexplored. One exception learns a dynamics model from human demonstrations and embeds it in MPC for stabilization \cite{belkhale2021model}, but targets near-static, vertically suspended loads, limiting dynamic maneuvering. Other high-performance results rely on precise estimation stacks and specialized hardware restricted to fixed racing courses, limiting generalization \cite{kaufmann2023champion}. Simulation frameworks offer task flexibility via modular rewards but typically require training a separate policy per task \cite{xu2024omnidrones}.

Robustness to hardware/parameter variation has been addressed by learning latent extrinsics such as mass or thrust \cite{zhang2023learning}, though these require an adaptation phase and are mostly validated near hover. Hybrid RL with adaptive control improves robustness and can track arbitrary references, but still assumes partially known dynamics and is shown in constrained scenarios \cite{huang2023datt}. These gaps motivate a unified, general-purpose learning-based controller serving as a single robust tracker across diverse quadrotor tasks—including payload manipulation—without task-specific design, switching, or retraining.


\begin{figure}[t]
    \centering
    \includegraphics[width=\linewidth]{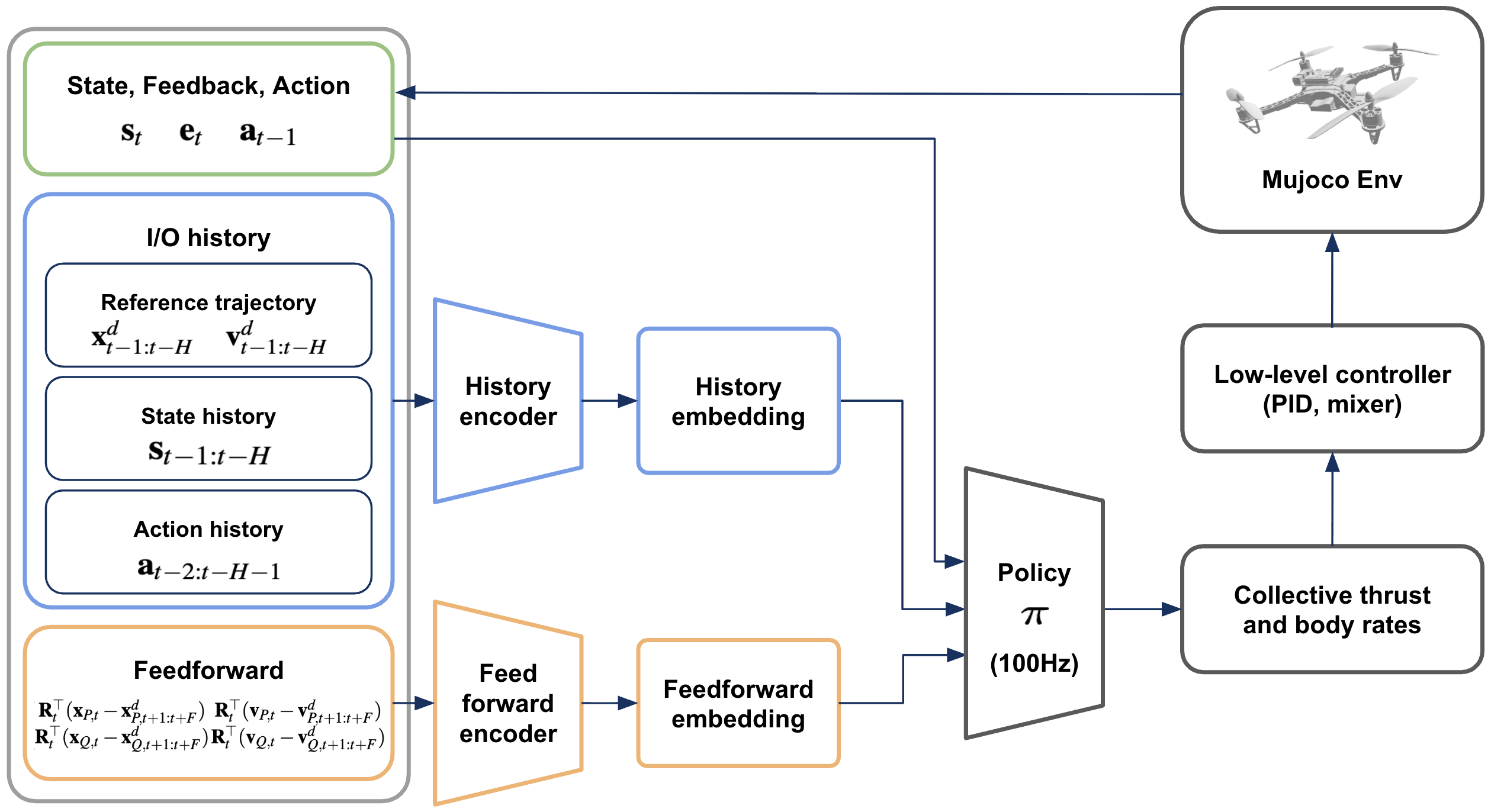}
    \caption{Diagram of the training loop. Present features are concatenated with history/preview embeddings and fed to $\pi$. Collective thrust and body rates (CTBR) commands pass through the rate loop and mixer before actuation.}
    \label{fig:algo_overview}
    \vspace{-4mm}
\end{figure}


\section{Problem Description}

\subsection{System Dynamics}

We consider two quadrotor variants: (1) a standard platform, and (2) a quadrotor with a flexible cable-suspended payload where cable length and payload mass may vary. Although configurations and observations differ, both admit the same control structure and can be handled within a unified learning-based framework.

The dynamics of the quadrotor-payload system are hybrid, with two operational modes: taut and slack. The active mode depends on the cable tension $T$, determined by the system state.

\subsubsection{Taut Mode Dynamics ($T>0$)}
When the cable is taut, the payload is constrained and the dynamics are tightly coupled. For a taut, massless cable, the system has eight degrees of freedom, configuration space $SE(3)\times S^2$, and four degrees of underactuation. The dynamics are \cite{sreenath2013geometric}:
\begin{align}
\mathbf{x}_Q &= \mathbf{x}_P - l\,\mathbf{q}, \\
\dot{\mathbf{x}}_P &= \mathbf{v}_P, \\
(m_Q + m_P)(\dot{\mathbf{v}}_P + g\,\mathbf{e}_3) 
&= \big(\mathbf{q}\!\cdot\! f\,\mathbf{R}\mathbf{e}_3 - m_Q l\,(\dot{\mathbf{q}}\!\cdot\!\dot{\mathbf{q}})\big)\,\mathbf{q} \notag\\
&\quad + \mathbf{w}_F, \\
\dot{\mathbf{q}} &= \boldsymbol{\omega} \times \mathbf{q}, \\
m_Q l\,\dot{\boldsymbol{\omega}} &= -\,\mathbf{q} \times f\,\mathbf{R}\mathbf{e}_3, \\
\dot{\mathbf{R}} &= \mathbf{R}\,\hat{\boldsymbol{\Omega}}, \label{eq:rot_kin} \\
\mathbf{J}_Q \dot{\boldsymbol{\Omega}} + \boldsymbol{\Omega} \times \mathbf{J}_Q \boldsymbol{\Omega} &= \mathbf{M} + \mathbf{w}_M. \label{eq:rot_dyn}
\end{align}
Here $\mathbf{x}_Q$ and $\mathbf{x}_P$ are the quadrotor and payload positions; $m_Q$, $m_P$ their masses; $\mathbf{q}\in S^2$ the cable direction; $l$ the cable length; $\mathbf{R}\in SO(3)$ the attitude; $f$ the total thrust; $\mathbf{M}$ the control moment; and $\mathbf{w}_F$, $\mathbf{w}_M$ external disturbances. Table~\ref{tab:quadrotor_payload_notation} summarizes the notation.

\subsubsection{Slack Mode Dynamics and Switching Conditions}
When the cable is slack ($T=0$, i.e., $\|\mathbf{x}_Q - \mathbf{x}_P\| < l$), quadrotor and payload dynamics decouple. The quadrotor behaves as a standard $SE(3)$ system governed by
\begin{equation}
    m_Q \ddot{\mathbf{x}}_Q = f \mathbf{R} \mathbf{e}_3 - m_Q g \mathbf{e}_3 + \mathbf{w}_F,
\end{equation}
while the payload acts as a free-falling point mass,
\begin{equation}
    m_P \ddot{\mathbf{x}}_P = -m_P g \mathbf{e}_3.
\end{equation}
The rotational dynamics remain as in Eq.~\eqref{eq:rot_kin}-\eqref{eq:rot_dyn}.

Mode transitions follow state-dependent guard conditions. Slack→taut occurs when the cable becomes taut ($\|\mathbf{x}_Q - \mathbf{x}_P\| \ge l$), representing a non-smooth impact. Taut→slack occurs when tension $T$ drops to zero.

This explicit hybrid structure, with non-smooth transitions, challenges traditional model-based control, which typically requires supervisory logic to detect the active mode and switch between laws. In contrast, our work learns a \emph{single, continuous} policy that acts as an implicit hybrid controller. It does not monitor switching conditions but learns a robust strategy from interaction data that navigates transitions. While the above analysis uses an idealized model for clarity, all simulations are conducted in MuJoCo with a finite-stiffness cable, ensuring the learned policy is validated on a realistic flexible model capturing slack–taut dynamics.


\subsection{Arbitrary Trajectory Tracking}

We pose a robust reference–tracking problem under dynamic, uncertain conditions. The reference $\mathbf{x}_\mathrm{ref}(t)$ is sampled from a family of sinusoid trajectories with randomized amplitudes, frequencies, and phases, and with hovering intervals at the start and end.
Trained on this distribution, the controller shows zero-shot generalization to unseen references from the same family. The task contains sharp turns and dynamically infeasible segments, where flatness-based controllers typically fail due to precomputed flat outputs.
Robustness is evaluated under two conditions: external disturbances, modeled as impulse forces in $[-0.5, 0.5]~\mathrm{N}$ and torques in $[-0.005, 0.005]~\mathrm{N\cdot m}$ applied for $[0, 0.5]~\mathrm{s}$; and initial-state perturbations, introduced as random offsets in position, orientation, and linear/angular velocity for both quadrotor and payload. Across cases, the controller rejects disturbances, stabilizes hover, and resumes accurate tracking, serving as a robust feedback tracker for both system configurations.


\section{End-to-End Model-Free RL for Quadrotor Payload Trajectory Tracking}

\subsection{Preliminaries}
We formulate the problem as a POMDP $ \mathcal{M} $, defined as a tuple $ (\mathcal{S}, \mathcal{A}, \mathcal{O}, \mathcal{T}, \gamma, r) $, where an agent interacts with an environment with state space $ \mathcal{S} $ and action space $ \mathcal{A} $, governed by transition function $ \mathcal{T}(\mathbf{s}_{t+1} | \mathbf{s}_t, \mathbf{a}_t) $. The agent receives observation $ \mathbf{o}_t \in \mathcal{O} $ through the observation function $ O(\mathbf{o}_t | \mathbf{s}_t, \mathbf{a}_{t-1}) $, which depends on the current state and previous action. Episodes start from an initial distribution $ \rho_0 $ over $\mathcal{S}$, i.e., $ \mathbf{s}_0 \sim \rho_0 $. Given reward $ r(\mathbf{o}_t, \mathbf{a}_t) $ and discount factor $ \gamma $, the objective in RL is to maximize the expected discounted return:
$ \mathbb{E} \left[ \sum_{t=0}^{\infty} \gamma^t r(\mathbf{o}_t, \mathbf{a}_t) \right] $.
The goal is to learn an optimal policy $ \pi^* \in \Pi $ mapping observations $ \mathbf{o}_t \in \mathcal{O} $ to optimal actions $ \mathbf{a}_t^* \in \mathcal{A} $.


\subsection{Algorithm and Training Overview}

We train a policy $\pi_\theta$ mapping observations to actions at $100\,\text{Hz}$. The observation includes \emph{present}, \emph{past}, and \emph{future} parts: present features $(\mathbf{s}_t,\mathbf{e}_t,\mathbf{a}_{t-1})$; a short I/O history $\boldsymbol{\zeta}_t$ of recent states, references, and actions over $H$ steps; and a feedforward term $\boldsymbol{\xi}_t$ with previewed position/velocity in the body frame. As in Fig.~\ref{fig:algo_overview}, the history and feedforward stacks pass through MLP encoders to form embeddings, concatenated with present features and fed to $\pi_\theta$. The actor and critic are separate two-layer MLPs (256$\times$256) with SiLU activations. The policy is trained with PPO \cite{schulman2017proximal}.

\begin{table}[t]
    \centering
    \footnotesize
    \setlength{\tabcolsep}{4pt}
    \renewcommand{\arraystretch}{1.1}
    \caption{Notation and definitions for the quadrotor–payload system.}
    \label{tab:quadrotor_payload_notation}
    \begin{tabularx}{\linewidth}{@{}L >{\RaggedRight\arraybackslash}X@{}}
        \toprule
        m_Q \in \mathbb{R} & Mass of the quadrotor \\
        \mathbf{J}_Q \in \mathbb{R}^{3\times 3} & Inertia matrix of the quadrotor w.r.t.\ the body-fixed frame \\
        \mathbf{R} \in SO(3) & Rotation matrix from the body-fixed frame to the inertial frame \\
        \boldsymbol{\Omega} \in \mathbb{R}^3 & Angular velocity of the quadrotor in the body-fixed frame \\
        \mathbf{x}_Q,\ \mathbf{v}_Q \in \mathbb{R}^3 & Position and velocity of the quadrotor CoM in the inertial frame \\
        \mathbf{x}_Q^d,\ \mathbf{v}_Q^d \in \mathbb{R}^3 & Desired quadrotor position and velocity from the reference trajectory \\
        f \in \mathbb{R} & Magnitude of the total thrust \\
        \mathbf{M} \in \mathbb{R}^3 & Body-frame moment \\
        \mathbf{w}_F \in \mathbb{R}^3 & External force disturbance on the system CoM \\
        \mathbf{w}_M \in \mathbb{R}^3 & External moment disturbance applied to the quadrotor \\
        m_P \in \mathbb{R} & Mass of the suspended load \\
        \mathbf{q} \in S^2 \subset \mathbb{R}^3 & Unit vector from quadrotor to load (cable direction) \\
        \boldsymbol{\omega} \in \mathbb{R}^3 & Angular velocity of the suspended load \\
        \mathbf{x}_P,\ \mathbf{v}_P \in \mathbb{R}^3 & Position and velocity of the suspended load (inertial frame) \\
        \mathbf{x}_P^d,\ \mathbf{v}_P^d \in \mathbb{R}^3 & Desired payload position and velocity from the reference trajectory \\
        l \in \mathbb{R} & Cable length \\
        T \in \mathbb{R} & Cable tension magnitude \\
        \bar{f} \in \mathbb{R} & Max total thrust \\
        \bar{\Omega} \in \mathbb{R} & Max body rate \\
        \bar{M},\ \underline{M} \in \mathbb{R} & Max and min body-frame moment \\
        \bottomrule
    \end{tabularx}
\end{table}


\subsection{Belief-Based Analysis of I/O History for Implicit Hybrid Control}

The hybrid nature of the quadrotor–payload system presents a significant challenge in partial observability. The discrete operational mode—taut or slack—is a latent variable that is not directly measured and dictates the system's underlying dynamics. Inferring this mode and the full continuous state $ \mathbf{s}_t $ from a single observation $ \mathbf{o}_t $ is intractable. To address this, our policy uses a fixed-length history of observations and actions (\textit{I/O history}). This does not violate the Markov property but rather restores it by providing a practical surrogate for the true belief state over both the discrete mode and continuous state. This history allows the policy to function as an implicit observer, learning to infer the system's hybrid state and enabling the stable, continuous control actions characteristic of an implicit hybrid controller.


\subsubsection{Belief-Based Policy and Value under Uncertainty}
In partially observable settings, the agent maintains a belief over the latent state, updated via Bayesian filtering:
\begin{equation}
\resizebox{\linewidth}{!}{$
\begin{aligned}
b_t(\mathbf{s}_t)
&= p\!\left(\mathbf{s}_t \mid \mathbf{o}_{0:t},\, \mathbf{a}_{0:t-1}\right) \\
&= \frac{ O\!\left(\mathbf{o}_t \mid \mathbf{s}_t,\, \mathbf{a}_{t-1}\right) }
        { p\!\left(\mathbf{o}_t \mid \mathbf{o}_{0:t-1},\, \mathbf{a}_{0:t-1}\right) }
    \sum_{\mathbf{s}_{t-1}} T\!\left(\mathbf{s}_t \mid \mathbf{s}_{t-1},\, \mathbf{a}_{t-1}\right)\,
    b_{t-1}\!\left(\mathbf{s}_{t-1}\right)
\end{aligned}
$}
\end{equation}
This update reduces entropy by exploiting temporal context:
\begin{equation}
H(\mathbf{s}_t \mid \mathbf{o}_{0:t}, \mathbf{a}_{0:t-1}) \leq H(\mathbf{s}_t \mid \mathbf{o}_t).
\end{equation}
Lower-entropy beliefs improve performance under:
\begin{equation}
\resizebox{\linewidth}{!}{$
V^*(b_t) = \max_{\mathbf{a}_t} \left[ 
\sum_{\mathbf{s}_t} b_t(\mathbf{s}_t) R(\mathbf{s}_t, \mathbf{a}_t)
+ \gamma \sum_{\mathbf{o}_{t+1}} p(\mathbf{o}_{t+1} \mid b_t, \mathbf{a}_t) V^*(b_{t+1}) \right],
$}
\end{equation}
and satisfy $ V^*(b_t) \geq V^*(b'_t) $ for any less informative belief $ b'_t(\mathbf{s}_t) = p(\mathbf{s}_t \mid \mathbf{o}_t) $. In actor–critic terms, concentrated beliefs yield sharper value estimates:
\begin{align}
b_t(\mathbf{s}_t)&\approx \delta(\mathbf{s}_t-\mathbf{s}^*_t)
\Rightarrow V^\pi(b_t)\approx \sum_{\mathbf{a}_t}\pi(\mathbf{a}_t\mid b_t)\,Q^\pi(\mathbf{s}^*_t,\mathbf{a}_t),\\
b_t(\mathbf{s}_t)&\approx \mathcal{U}(S)
\Rightarrow V^\pi(b_t)\approx \frac{1}{|\mathcal S|}\sum_{\mathbf{s}_t,\mathbf{a}_t}\pi(\mathbf{a}_t\mid b_t)\,Q^\pi(\mathbf{s}_t,\mathbf{a}_t).
\end{align}
Hence, belief accuracy is critical in POMDPs.


\subsubsection{Hybrid Dynamics and Implicit Mode Estimation}
The core challenge for our I/O history-based approach is estimating the unobserved hybrid mode $m_t$. For the flexible cable system, dynamics can be modeled as:
\begin{equation}
\Sigma = 
\begin{cases}
\mathbf{s}_{t+1} = f_m(\mathbf{s}_t, \mathbf{a}_t), & m \in \{\text{taut}, \text{slack}\}, \\
\mathbf{s}_{t+1}^+ = \Delta_{m^- \rightarrow m^+}(\mathbf{s}_{t}^-), & \mathbf{s}_t \in S_\text{switch},
\end{cases}
\end{equation}
where mode $ m_t $ is unobserved. An accurate belief over the mode is critical, as the optimal action can differ drastically between taut and slack regimes. The agent must maintain a joint belief over mode and state:
\begin{equation}
b_t(m_t, \mathbf{s}_t) = p(m_t, \mathbf{s}_t \mid \mathbf{o}_{0:t}, \mathbf{a}_{0:t-1}).
\end{equation}
Updating this belief requires marginalizing over past modes and states:
\begin{equation}
\begin{aligned}
p(m_t \mid \mathbf{o}_{0:t}, \mathbf{a}_{0:t-1}) 
&\propto \sum_{m_{t-1}} \int p(m_t \mid m_{t-1}, \mathbf{s}_{t-1}, \mathbf{a}_{t-1}) \\
&\qquad \times b_{t-1}(m_{t-1}, \mathbf{s}_{t-1})\, d\mathbf{s}_{t-1}.
\end{aligned}
\end{equation}
Because mode transitions depend on latent variables (like cable tension), a single observation $ \mathbf{o}_t $ is insufficient; temporal context is needed to disambiguate modes. Entropy shrinks with longer histories:
\begin{equation}
\begin{aligned}
H(m_t,\mathbf{s}_t \mid \mathbf{o}_t)
&> H(m_t,\mathbf{s}_t \mid \mathbf{o}_{t-1},\mathbf{o}_t,\mathbf{a}_{t-1}) \\
&> H(m_t,\mathbf{s}_t \mid \mathbf{o}_{0:t},\mathbf{a}_{0:t-1}).
\end{aligned}
\end{equation}
This grounding shows why I/O history is the key mechanism enabling our policy to act as an implicit hybrid controller. By providing temporal context, the history lets the policy form a richer internal representation of the system's belief state, reducing uncertainty over both discrete mode ($m_t$) and continuous state ($\mathbf{s}_t$). This ability to implicitly infer hybrid dynamics without an explicit observer or switching logic allows a single, continuous policy to achieve robust control across non-smooth transitions. The importance of this mechanism is empirically validated by our ablation studies in Table~\ref{tab:ablation-history}.


\subsection{State, Action, and Observation Spaces}

\subsubsection{State}

We define the state as \begin{equation} \mathbf{s}_t = \begin{bmatrix} \mathbf{x}_{Q,t} & \mathbf{R}_t & \mathbf{x}_{P,t} & \mathbf{v}_{Q,t} \\ \boldsymbol{\omega}_{Q,t} & \mathbf{v}_{P,t} & m_{P} & l \end{bmatrix}. \end{equation} Position and velocity vectors are normalized element-wise by their respective maxima $x_Q^{\max}$, $x_P^{\max}$, $v_Q^{\max}$, and $v_P^{\max}$. Attitude is represented by a rotation matrix to avoid singularities and remains compatible with geometric-control formulations. Unlike existing work that treats payload configuration as unknown parameters and only ensures policy performance within a prior parameter set, we explicitly include payload information by augmenting mass and cable length as states. 

\subsubsection{Action} We adopt a collective thrust and body rates (CTBR) parameterization and convert it to individual rotor thrusts via an inner rate loop and a static mixer \cite{kaufmann2022benchmark}. At time $t$, the policy outputs a normalized action $\mathbf{a}_t=\big[a_{c,t}\; a_{p,t}\; a_{q,t}\; a_{r,t}\big]\in[-1,1]^4$, which we map to the total thrust and desired body-rate setpoints as \begin{equation} f_t=\tfrac{\bar f}{2}\big(1+a_{c,t}\big),\qquad \boldsymbol{\Omega}^{\,d}_t=\bar{\Omega}\,\big[a_{p,t}\;a_{q,t}\;a_{r,t}\big]. \end{equation} A smooth squashing (e.g., $\tanh$) may be used; in practice we use linear rate scaling and an affine map for $f_t$. The desired rates $\boldsymbol{\Omega}^{\,d}_t$ are tracked by a per-axis PID acting on $\boldsymbol{\Omega}_t$ with anti-windup and saturation, yielding $\mathbf{M}_t\in\mathbb{R}^3$; this inner loop decouples attitude from thrust and is robust to delays and model mismatch. Finally, $(f_t,\mathbf{M}_t)$ is mapped to single-rotor thrusts via a fixed mixer $\mathbf{A}\in\mathbb{R}^{4\times 4}$ set by geometry and rotor coefficients, \begin{equation} \mathbf{f}_t=\operatorname{clip}\!\left(\mathbf{A}\,[\,f_t,\ \mathbf{M}_t],\ 0,\ \bar{f}\right), \end{equation} and passed through first-order motor dynamics. To reduce sim-to-real gap, we also inject a 10--30\,ms discrete input delay at 100\,Hz and apply a first-order rotor lag with asymmetric rise/fall time constants, randomized during training. 


\subsubsection{Observation}

The observation captures the current state, short I/O history, and a preview of the reference, providing context for robust decisions. It is structured as \begin{equation} \mathbf{o}_t = \begin{bmatrix} \mathbf{s}_t & \mathbf{e}_t & \mathbf{a}_{t-1} & \mathbf{\zeta}_t & \mathbf{\xi}_t \end{bmatrix}, \end{equation} where $\mathbf{s}_t$ is the current state and $\mathbf{a}_{t-1}$ the previous action. $\mathbf{e}_t \in \mathbb{R}^{12}$ is the body-frame tracking error: \begin{equation} \mathbf{e}_t = \begin{bmatrix} \mathbf{R}_t^\top \mathbf{e}_{x_{P, t}} & \mathbf{R}_t^\top \mathbf{e}_{v_{P, t}} & \mathbf{R}_t^\top \mathbf{e}_{x_{Q, t}} & \mathbf{R}_t^\top \mathbf{e}_{v_{Q, t}} \end{bmatrix}, \end{equation} where $ \mathbf{e}_{x_P}, \mathbf{e}_{v_P}, \mathbf{e}_{x_Q}, \mathbf{e}_{v_Q}$ are the position/velocity errors for payload and quadrotor. $\mathbf{\zeta}_t$ is the I/O history: \begin{equation} \mathbf{\zeta}_t = \begin{bmatrix} \mathbf{s}_{t-1:t-H} & \mathbf{x}_{t-1:t-H}^d & \mathbf{v}_{t-1:t-H}^d & \mathbf{a}_{t-2:t-H-1} \end{bmatrix}, \end{equation} with buffer length $H{=}5$ and \begin{equation} \begin{aligned} \mathbf{x}_{t-1:t-H}^d &= \big[\,\mathbf{x}_{Q,\,t-1:t-H}^d\;\;\mathbf{x}_{P,\,t-1:t-H}^d\,\big],\\ \mathbf{v}_{t-1:t-H}^d &= \big[\,\mathbf{v}_{Q,\,t-1:t-H}^d\;\;\mathbf{v}_{P,\,t-1:t-H}^d\,\big]. \end{aligned} \end{equation} $\mathbf{\xi}_t$ is the body-frame feedforward term w.r.t. the preview: 
\begin{equation} \mathbf{\xi}_t = \begin{bmatrix} \mathbf{R}_t^\top (\mathbf{x}_{P, t} - \mathbf{x}_{P, t+1:t+F}^d) & \mathbf{R}_t^\top (\mathbf{v}_{P, t} - \mathbf{v}_{P, t+1:t+F}^d) \\ \mathbf{R}_t^\top (\mathbf{x}_{Q, t} - \mathbf{x}_{Q, t+1:t+F}^d) & \mathbf{R}_t^\top (\mathbf{v}_{Q, t} - \mathbf{v}_{Q, t+1:t+F}^d) \end{bmatrix} \end{equation} \noindent To emulate sensor imperfections and promote robustness under partial observability, we inject clipped Gaussian noise during training: positions $\mathbf{x}_Q,\mathbf{x}_P$ use $\sigma_x{=}0.01$\,m with $\pm 2.5$\,mm clipping; linear velocities $\mathbf{v}_Q,\mathbf{v}_P$ use $\sigma_v{=}0.02$\,m/s with $\pm 5$\,mm/s clipping; attitude uses $\tilde{\mathbf{R}}=\mathbf{R}(\boldsymbol{\theta}+\eta_\theta)$ with $\sigma_\theta{=}\pi/60$\,rad and $\pm\pi/120$\,rad clipping; body rates $\boldsymbol{\Omega}$ use $\sigma_\Omega{=}\pi/30$\,rad/s with $\pm\pi/60$\,rad/s clipping. In principle, a full history would be ideal but is intractable; a fixed-length buffer offers a practical POMDP approximation that preserves essential temporal information \cite{meuleau2013learning}. While I/O history is common in legged locomotion \cite{li2024reinforcement, peng2018sim}, high control rates in aerial robots (often $>50$ Hz) make compact encodings important. With a payload, the hybrid and more underactuated dynamics further increase the value of history: it aids implicit dynamics learning (analogous to model-based RL that first learns dynamics, then optimizes control) and provides temporal feedback that improves robustness to delays and sensor noise \cite{li2024reinforcement}. Finally, organizing the observation into \emph{present} (feedback/state/action), \emph{past} (I/O history), and \emph{future} (feedforward preview) mirrors finite-horizon optimal-control structure and helps the policy exploit known future references when maximizing return \cite{bertsekas2012dynamic}.


\subsection{Rewards and Episode Design}

\subsubsection{Reward structure}

We balance tracking accuracy with stable, smooth control by summing exponentiated penalties on key quantities (position, yaw, body rates, cable motion, action magnitude, and action variation): \begin{equation} \begin{aligned} r_{x_P} &= w_{x_P}\,\exp\!\big(-\alpha_{x_P}\,\lVert \mathbf{x}_{P,t}-\mathbf{x}^d_{P,t}\rVert\big),\\ r_{\psi} &= w_{\psi}\,\exp\!\big(-\alpha_{\psi}\,|\psi_t|\big),\\ r_{\Omega} &= w_{\Omega}\,\exp\!\big(-\alpha_{\Omega}\,\lVert \boldsymbol{\Omega}_t\rVert\big),\\ r_{\dot q} &= w_{\dot q}\,\exp\!\big(-\alpha_{\dot q}\,\lVert \dot{\mathbf{q}}_t\rVert\big),\\ r_{a} &= w_{a}\,\exp\!\big(-\alpha_{a}\,\lVert \mathbf{a}_t\rVert\big),\\ r_{\Delta a} &= w_{\Delta a}\,\exp\!\big(-\alpha_{\Delta a}\,\lVert \Delta\mathbf{a}_t\rVert\big). \end{aligned} \end{equation} Here $\psi_t$ is the yaw angle, $\dot{\mathbf{q}}_t$ is the rate of the cable direction (approximated via relative quadrotor–payload velocity normalized by cable length), $\Delta\mathbf{a}_t$ is an exponentially weighted $L_2$ norm of recent action changes, and $\alpha_k,w_k$ are scale and weight coefficients. The total reward is \begin{equation} r_t = r_{x_P} \left(1 + r_{\psi} + r_{\Omega} + r_{\dot{q}} \right) + r_a + r_{\Delta a}. \end{equation} This structure prioritizes accurate payload tracking and further rewards low yaw, low body rates, and small cable motion, while discouraging large or rapidly changing actions. \subsubsection{Early termination} For sample efficiency, episodes terminate on failure: ground contact ($z_Q<0$), excessive payload error $\lVert \mathbf{x}_{P,t}-\mathbf{x}_{P,t}^d\rVert>\varepsilon_{\text{pos}}$ or $\lVert \mathbf{v}_{P,t}-\mathbf{v}_{P,t}^d\rVert>\varepsilon_{\text{vel}}$, or if any quadrotor Euler angle exceeds $\pi/2$ in magnitude.


\subsection{Control Perspective and Stability Considerations}
At a high level, the learned policy issues CTBR commands while a faster rate–PID and static mixer produce body moments and single–rotor thrusts; with anti–windup and saturation, the inner loop behaves nearly as an identity from $\boldsymbol{\Omega}^{\,d}$ to body rates.

\textit{Robustness view.} Around hover, the inner loop stabilizes the rotation while the outer loop acts on the translational/cable dynamics through thrust direction and magnitude. With bounded delays and first–order rotor lags, a small–gain/ISS argument implies bounded tracking error provided: (i) the rate loop remains stable with margin under saturation; (ii) the policy outputs are bounded (by design); and (iii) disturbances enter as additive inputs. Domain randomization effectively enlarges these margins by training across variability in $(m_P,l)$ and aerodynamics.

\textit{Energy/damping view.} The reward terms on body rates and cable motion ($r_{\Omega}$, $r_{\dot q}$) act as damping injection for the coupled quadrotor–pendulum modes, while CTBR allows smooth thrust–direction shaping. This encourages thrust alignment with the cable tension and attenuates swing without an explicit model.


\begin{figure}[t]
    \centering
    \begin{subfigure}[t]{0.49\linewidth}
        \centering
        \includegraphics[width=\linewidth]{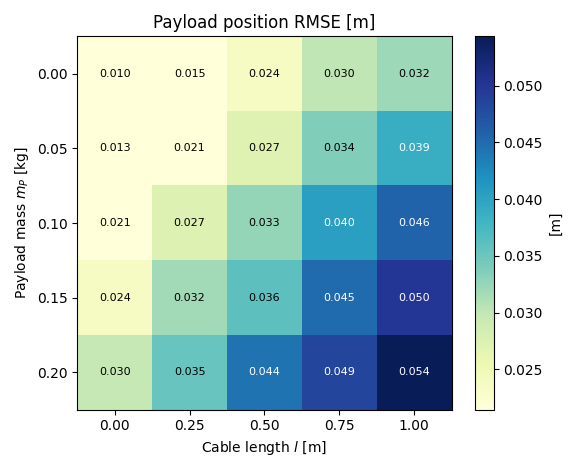}
        \caption{Payload position RMSE.}
    \end{subfigure}\hfill
    \begin{subfigure}[t]{0.49\linewidth}
        \centering
        \includegraphics[width=\linewidth]{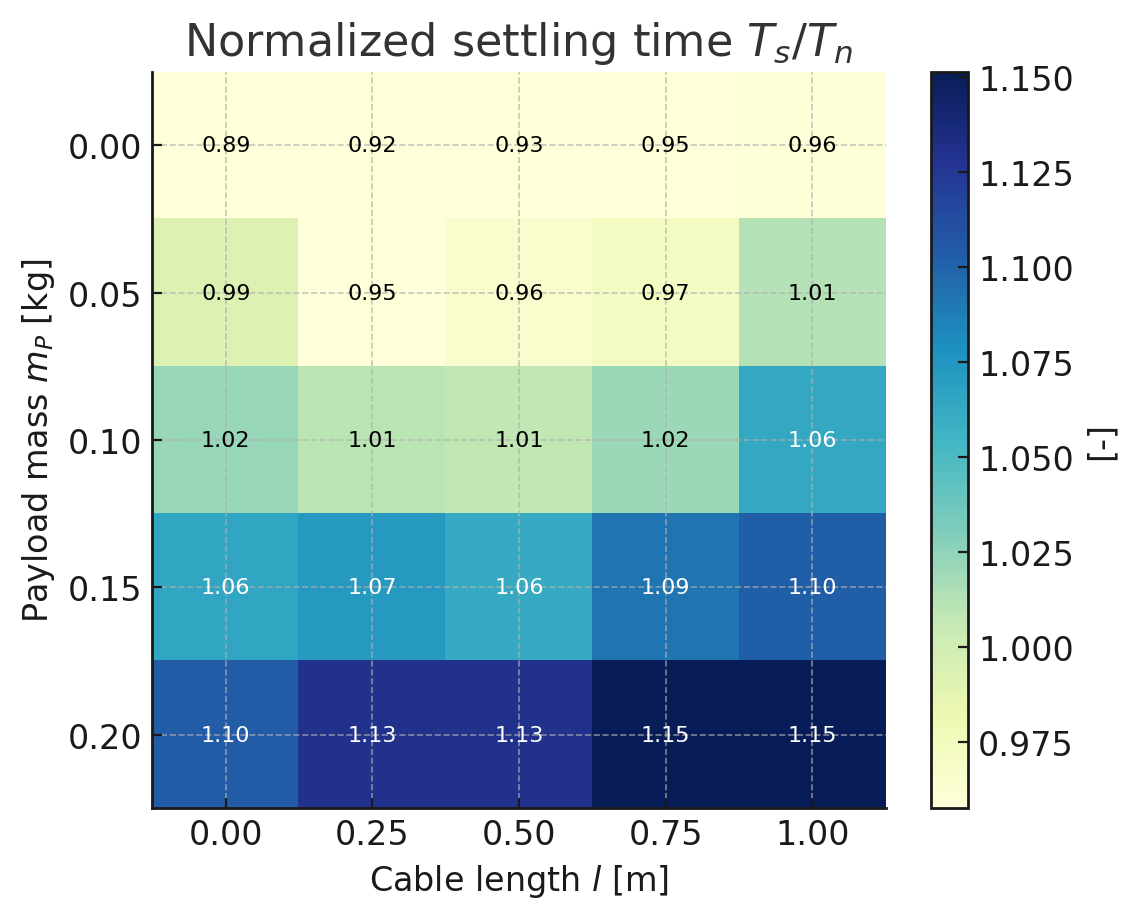}
        \caption{Normalized settling time.}
    \end{subfigure}
    \caption{Zero-shot performance across payload mass $m_P$ and cable length $l$ on unseen trajectories.}
    \label{fig:heatmaps}
\end{figure}


\subsection{Domain Randomization}

To promote robustness and generalization, we apply domain randomization at the start of each training episode \cite{tobin2017domain}. For the quadrotor, we perturb body mass, inertia, center of mass, and the orientation of the local inertia frame (via randomized translations and quaternion rotations); actuator inconsistencies are modeled by varying motor gear coefficients. For the suspended payload, mass and cable length are uniformly sampled from $[0.0, 0.2]~\text{kg}$ and $[0.0, 1.0]~\text{m}$, respectively, and the payload pose and cable constraints are updated accordingly. We also randomize near-ground aerodynamic effects: when $z<0.5~\text{m}$, a force with direction sampled from the upper hemisphere is applied, with magnitude increasing as altitude decreases. Jointly randomizing these physical and aerodynamic properties exposes the policy to cases ranging from no payload to strong disturbances, yielding zero-shot generalization without controller switching or environment-specific tuning. Empirical performance across the randomized configuration grid is summarized in the heatmaps of Fig.~\ref{fig:heatmaps}.


\section{Numerical Results}

We evaluate \textsc{RoVerFly} on trajectory tracking and hover stabilization in both \emph{quadrotor-only} and \emph{flexible cable–suspended payload} configurations. Unless noted otherwise, a \emph{single} end-to-end policy (no controller switching or re-tuning) is used for all conditions. The policy runs at $100$~Hz and the simulator at $500$~Hz. During training we inject sensor noise, an actuation delay of 10--30~ms, and parameter randomization. Quadrotor parameters are summarized in Table~\ref{tab:quadrotor_param}.


\begin{figure}[t]
    \centering
    \begin{subfigure}[t]{0.95\linewidth}
        \centering
        \includegraphics[width=\linewidth]{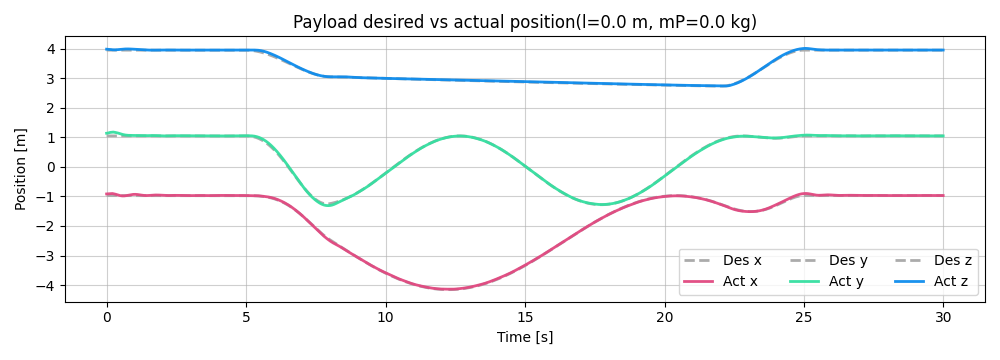}
    \end{subfigure}

    \vspace{0.5em}

    \begin{subfigure}[t]{0.95\linewidth}
        \centering
        \includegraphics[width=\linewidth]{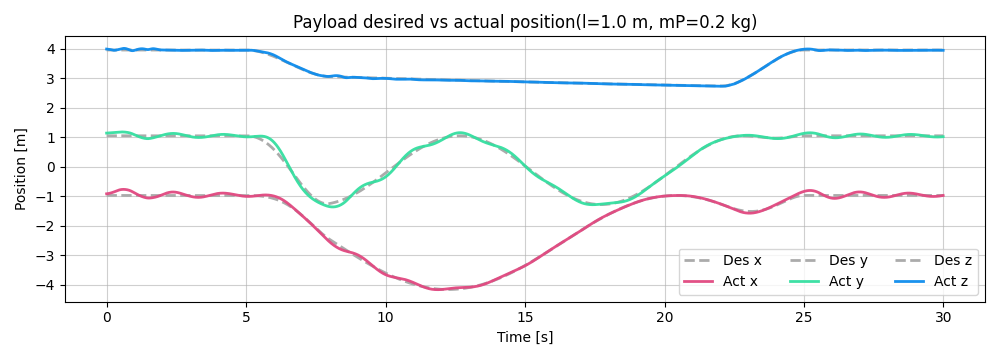}
    \end{subfigure}
    \caption{Payload tracking on unseen trajectories: without payload and with flexible cable-suspended payload.}
    \label{fig:xp-tracking}
    \vspace{-4mm}
\end{figure}


\begin{table}[t]
    \centering
    \begin{tabular}{l|c}
    Parameters & Values \\
    \hline
    $m_Q$ & $0.835~\text{kg}$ \\
    $m_P$ & $0.0-0.2~\text{kg}$ \\
    $J_Q$ & $Diag([4.01, 3.58, 6.36]) \times 10^{-3}~\text{kg}\cdot\text{m}^2$ \\
    $\bar{f}$ & $30~\text{N}$ \\
    
    \end{tabular}
    \vspace{0.6em}
    \caption{Nominal quadrotor parameters.}
    \label{tab:quadrotor_param}
    \vspace{-4mm}
\end{table}

\subsection{Tasks and Evaluation Protocol}
We consider two tasks: (i) \textbf{arbitrary trajectory tracking} under unseen references and disturbances, and (ii) \textbf{hover stabilization} under large initial perturbations. For tracking, reference families are generated by summing sinusoids with randomized amplitudes, frequencies, and phases; evaluation uses held-out trajectories (zero-shot). For hover, we apply a randomized initial state perturbation and measure recovery.

For tracking, we report per-axis position RMSE and the time-averaged RMSE norm. For hover, we use the settling-time metric
$T_s=\min\{t:\lvert e(\xi)\rvert\le\varepsilon,~\forall \xi\in[t,t+\tau]\}$,
with $\varepsilon=0.01$~m and $\tau=0.5$~s, the normalized $T_s/T_n$ with $T_n=2\pi\sqrt{l/g}$ (cable–pendulum natural period), and the steady-state error $e_{ss}$ computed as the mean error over the last $0.5$~s. All results aggregate multiple random seeds and unseen trajectories.


\subsection{Zero-Shot Tracking Across Payload Configurations}
We evaluate the same policy at two endpoints of the domain: \emph{without payload} $(l{=}0.0~\mathrm{m},~m_P{=}0.0~\mathrm{kg})$ and \emph{with flexible payload} $(l{=}1.0~\mathrm{m},~m_P{=}0.2~\mathrm{kg})$. The controller tracks the desired payload CoM in $x,y,z$ without any controller switching. Quantitatively, without a payload the RMSE is $x=0.008~\mathrm{m},~y=0.010~\mathrm{m},~z=0.008~\mathrm{m}$ (total $0.016~\mathrm{m}$); with a flexible cable-suspended payload it is $x=0.024~\mathrm{m},~y=0.010~\mathrm{m},~z=0.006~\mathrm{m}$ (total $0.034~\mathrm{m}$). The policy remains stable through hover segments and aggressive turns, illustrating strong zero-shot generalization across dynamics.


\subsection{Hover Stabilization and Disturbance Rejection}
For hover, we sample an initial perturbation as 
$\Delta \mathbf{x}_Q\!\sim\! \mathcal{U}([-0.1,0.1]^3)\,\mathrm{m}$ applied to both quadrotor and payload positions, 
$\Delta \mathbf{v}_Q\!\sim\! \mathcal{U}([-0.1,0.1]^3)\,\mathrm{m/s}$ to both linear velocities, 
$\Delta \mathbf{\theta}_Q\!\sim\! \mathcal{U}([-\pi/12,\pi/12]^3)\,\mathrm{rad}$ to the quadrotor attitude, and 
$\Delta \mathbf{\Omega}\!\sim\! \mathcal{U}([-\pi/12,\pi/12]^3)\,\mathrm{rad/s}$ to the quadrotor body rates. 
The payload error responses $(e_x,e_y,e_z)$ remain within a small neighborhood and converge with $e_{ss}\!\approx\! 0$. For $l{=}1.0$~m, the observed settling time is on the order of a single natural period ($T_n\!\approx\!2.0$~s), indicating physically consistent recovery under underactuated cable–pendulum dynamics.

\begin{figure}[t]
    \centering
    \includegraphics[width=0.95\linewidth]{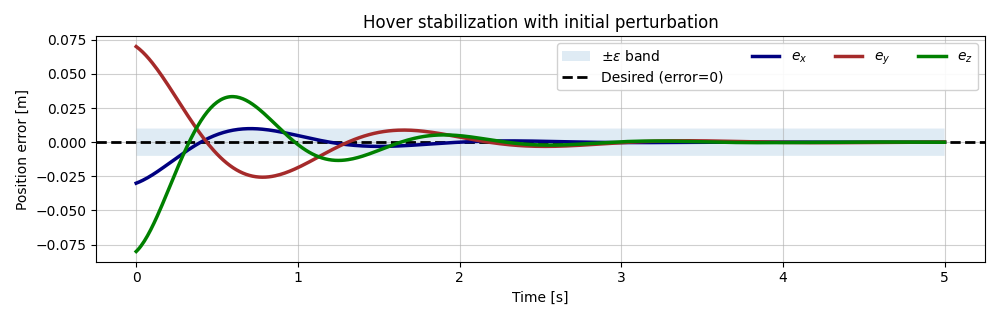}
    \caption{Hover initial perturbation rejection: position error traces ($l{=}1.0~\mathrm{m},~m_P{=}0.2~\mathrm{kg}$).}
    \label{fig:stabilizing}
    \vspace{-2.5mm}
\end{figure}


\subsection{Generalization Over Payload Mass and Cable Length}
We assess zero-shot generalization over a grid $m_P\!\in\![0,0.2]$~kg and $l\!\in\![0,1]$~m using unseen trajectories and multiple seeds per cell. Figure~\ref{fig:heatmaps} reports payload-position RMSE (tracking accuracy) and normalized settling time $T_s/T_n$ (recovery speed relative to $T_n$). RMSE grows with heavier payloads and longer cables, reflecting increased inertia and underactuation rather than brittle failures. Most $T_s/T_n$ values lie near $1$–$1.3$, indicating recovery within about one natural period. Short cables ($l\!\approx\!0$) yield the fastest recovery and smallest errors; larger $l$ introduces more oscillation yet remains well damped. Increasing $m_P$ slows the response but does not destabilize the policy. These trends indicate that domain/task randomization together with the history- and feedforward-augmented observation support a single policy that generalizes across $(m_P,l)$ without task-specific design or re-tuning.


\subsection{Ablation on I/O History}
We quantify the effect of temporal context (the last $H$ steps of state/action/reference). Using the same trained policy and evaluation setup, we vary $H\in\{0,1,5,10\}$ on unseen trajectories for two configurations: no payload $(l{=}0,\,m_P{=}0)$ and a flexible payload $(l{=}1.0~\mathrm{m},\,m_P{=}0.2~\mathrm{kg})$. Table~\ref{tab:ablation-history} reports payload position RMSE (per-axis and total).

Temporal history helps in both cases but is markedly more important with a flexible cable: removing history increases total RMSE from $0.054$ to $0.075$ ($+40\%$) with payload, versus $0.010$ to $0.021$ ($+110\%$) without payload. A minimal buffer ($H{=}1$) recovers part of the loss, a short window ($H{=}5$) is near-optimal under hybrid (taut/slack) dynamics and delay, and $H{=}10$ gives only marginal gains.


\begin{table}[t]
    \centering
    \footnotesize
    \setlength{\tabcolsep}{6pt}
    \renewcommand{\arraystretch}{1.12}
    \caption{Effect of I/O history length $H$ on payload position RMSE (m) for unseen trajectories.}
    \label{tab:ablation-history}
    \begin{tabular}{@{}lcccccc@{}}
        \toprule
        Configuration & H & $\mathrm{RMSE}_x$ & $\mathrm{RMSE}_y$ & $\mathrm{RMSE}_z$ & \textbf{Total} \\
        \midrule
        \multirow{4}{*}{$l{=}0,\,m_P{=}0$} 
            & 0  & 0.011 & 0.014 & 0.010 & \textbf{0.021} \\
            & 1  & 0.010 & 0.012 & 0.009 & \textbf{0.018} \\
            & 5  & 0.008 & 0.010 & 0.008 & \textbf{0.016} \\
            & 10 & 0.007 & 0.006 & 0.003 & \textbf{0.010} \\
        \addlinespace[2pt]
        \multirow{4}{*}{$l{=}1.0,\,m_P{=}0.2$} 
            & 0  & 0.051 & 0.047 & 0.031 & \textbf{0.075} \\
            & 1  & 0.049 & 0.043 & 0.029 & \textbf{0.072} \\
            & 5  & 0.045 & 0.041 & 0.025 & \textbf{0.066} \\
            & 10 & 0.038 & 0.032 & 0.020 & \textbf{0.054} \\
        \bottomrule
    \end{tabular}
    \vspace{-2.5mm}
\end{table}


\subsection{Reference Trajectory Generation}

We generate a family of \emph{payload-centric} references that combine (i) explicit hover segments and (ii) smooth oscillatory motion with randomized amplitudes and frequencies. Each episode draws
\begin{flalign}
& (a_x,a_y)\sim \mathcal{U}([-2,2]),\; a_z\sim \mathcal{U}([-1,1]), &&\\
& (f_1,f_2)\sim \mathcal{U}([-0.2,0.2]),\; f_3\sim \mathcal{U}([-0.1,0.1]). &&
\end{flalign}
with fixed phases $\phi_x,\phi_y,\phi_z\in\{\tfrac{\pi}{2},\tfrac{3\pi}{2}\}$. Let $\omega_i=2\pi f_i$ and $t_f$ be the episode horizon. Hover at the beginning and end is enforced by a smooth window $w(t)\in[0,1]$:
\begin{equation}
w(t)=
\begin{cases}
0, & t<t_s \text{ or } t>t_e,\\
3\alpha^2-2\alpha^3, & t\in[t_s,t_s+\Delta],\ \alpha=\dfrac{t-t_s}{\Delta},\\
1, & t\in[t_s+\Delta,\,t_e-\Delta],\\
3\beta^2-2\beta^3, & t\in[t_e-\Delta,t_e],\ \beta=\dfrac{t_e-t}{\Delta},
\end{cases}
\end{equation}
with $t_s=5~\mathrm{s}$, $t_e=t_f-5~\mathrm{s}$, and transition duration $\Delta=3~\mathrm{s}$. Its derivative $\dot w(t)$ avoids velocity/acceleration discontinuities at the transitions.

\subsubsection{Payload reference}
The desired payload position, velocity, and acceleration are
\begin{equation}
\begin{aligned}
\mathbf{x}_P^d(t) &=
\begin{bmatrix}
w(t)\,A_x\!\big(1-\cos(\omega_1 t+\phi_x)\big)\\
w(t)\,A_y\!\big(1-\cos(\omega_2 t+\phi_y)\big)\\
w(t)\,A_z\!\big(1-\cos(\omega_3 t+\phi_z)\big)
\end{bmatrix},\\[2pt]
\mathbf{v}_P^d(t) &= \dot{\mathbf{x}}_P^d(t), \qquad
\mathbf{a}_P^d(t) = \ddot{\mathbf{x}}_P^d(t).
\end{aligned}
\end{equation}
We cap nominal peak speed by normalizing $(A_x,A_y,A_z)$ if $\max\{A_x|\omega_1|, A_y|\omega_2|, A_z|\omega_3|\}>v_{\max}$ with $v_{\max}=4~\mathrm{m/s}$, ensuring comparable difficulty across draws while still covering aggressive segments.

\subsubsection{Cable direction and quadrotor reference}
Given payload mass $m_P$ and gravity $\mathbf{g}=g\mathbf{e}_3$, define
\begin{equation}
\mathbf{T}_P(t) = -\,m_P\big(\mathbf{a}_P^d(t)+\mathbf{g}\big), \qquad
\mathbf{q}^d(t) = \frac{\mathbf{T}_P(t)}{\|\mathbf{T}_P(t)\|}.
\end{equation}
We obtain $\dot{\mathbf{q}}^d$ and $\ddot{\mathbf{q}}^d$ by differentiating $\mathbf{T}_P$. For cable length $l$, the quadrotor position reference is
\begin{equation}
\mathbf{x}_Q^d(t)=\mathbf{x}_P^d(t)-l\,\mathbf{q}^d(t),
\end{equation}
with velocities/accelerations derived analogously.

This design (i) enforces hover at the start/end for consistent stabilization evaluation, (ii) excites coupled translation–pendulum modes via randomized amplitudes/frequencies, and (iii) preserves smooth motion on/off through $w(t)$ and $\dot w(t)$. Randomization over $(A_x,A_y,A_z,f_1,f_2,f_3)$ yields a controlled yet diverse task distribution; zero-shot tests use held-out parameters and unseen trajectories from the same family to probe robustness across payload mass, cable length, and reference complexity.



\begin{figure}[t]
    \centering
    \begin{subfigure}[t]{0.48\linewidth}
        \centering
        \includegraphics[width=\linewidth]{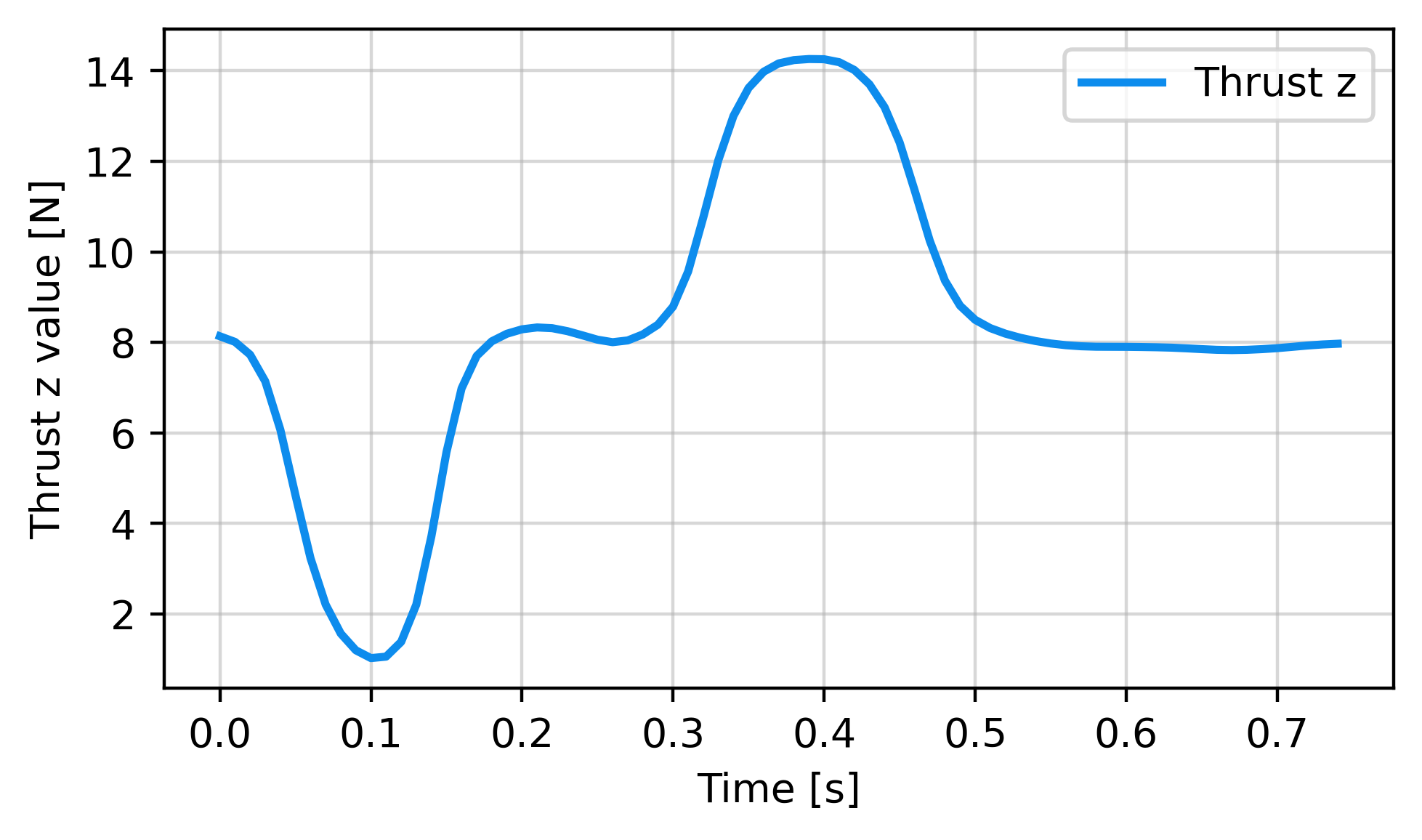}
        \caption{Quadrotor thrust in $z$ axis.}
        \label{fig:thrust_z}
    \end{subfigure}
    \hfill
    \begin{subfigure}[t]{0.48\linewidth}
        \centering
        \includegraphics[width=\linewidth]{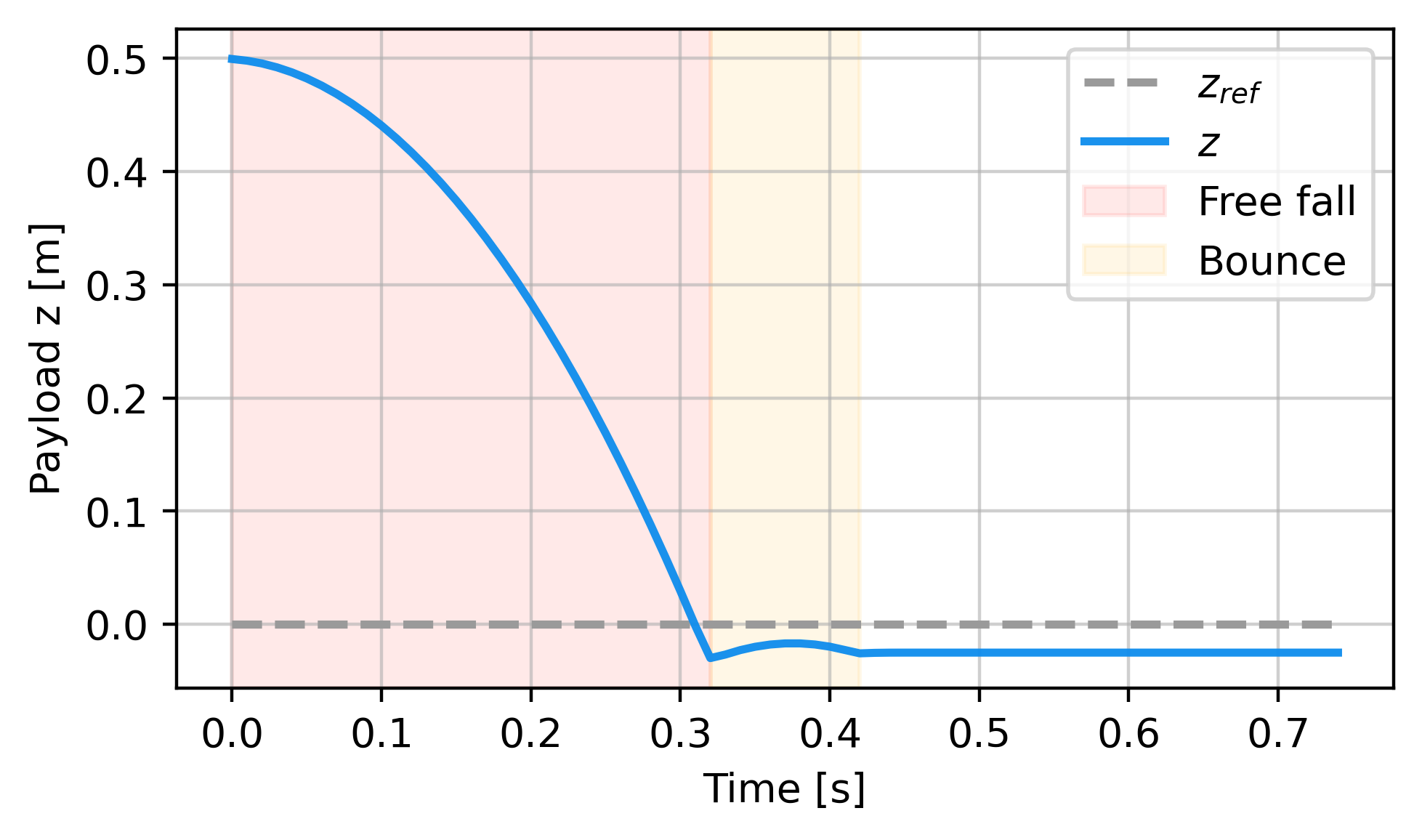}
        \caption{Payload trajectory in $z$.}
        \label{fig:payload_z}
    \end{subfigure}
    \caption{Controller behavior during slack-to-taut transition. 
    (a) The controller modulates thrust before tautening, mitigating impact. 
    (b) The payload exhibits distinct free-fall and bounce slack phases before settling. The shaded intervals highlight free-fall and bounce phases during slack-to-taut transition.}
    \label{fig:slack2taut_pair}
    \vspace{-4mm}
\end{figure}


\section{Conclusion and Future Work}

We presented \textsc{RoVerFly}, a unified learning-based controller that tracks arbitrary trajectories and stabilizes under disturbances across payload conditions, from no payload to flexible cable–suspended loads with varying mass and length. A single policy trained with task and domain randomization handles $(m_P,l)$ changes without switching or re-tuning. The observation design combines feedback, short I/O history, and feedforward terms, while CTBR parameterization yields smooth attitude shaping and effective payload damping. Experiments show (i) low tracking error across configurations, (ii) rapid disturbance rejection with near-zero steady-state error and settling near one pendulum period, (iii) smooth trends over $(m_P,l)$ without brittle failures, and (iv) ablations confirming the role of history and feedforward, indicating strong zero-shot generalization of a \emph{single} policy across hybrid, underactuated regimes.

Our analysis in uses the standard taut, massless cable model, while simulations employ a flexible cable with slack–taut transitions. The learned policy handles these empirically, but stability guarantees for the full hybrid system remain open. Evaluation is limited to high-fidelity simulation; unmodeled aerodynamics, and onboard compute limits are not yet tested. The policy also lacks formal safety or constraint guarantees. Future directions include hardware validation with varied sensors, rigorous benchmarking of slack–taut transitions, extension to contacts and time-varying cable length, testing on aggressive infeasible references, integration of safety wrappers (e.g., CBF shields) and online $(m_P,l)$ identification, exploration of meta-RL and few-shot adaptation for out-of-distribution payloads and environments, scaling to multi-UAV cooperative transport, and developing analytical links (e.g., input-to-state stability, robustness margins) between the learned policy and control guarantees.




\printbibliography

@inproceedings{lee2010geometric,
  title={Geometric tracking control of a quadrotor UAV on SE (3)},
  author={Lee, Taeyoung and Leok, Melvin and McClamroch, N Harris},
  booktitle={49th IEEE conference on decision and control (CDC)},
  pages={5420--5425},
  year={2010},
  organization={IEEE}
}

@INPROCEEDINGS{sreenath2013geometric,
  author={Sreenath, Koushil and Lee, Taeyoung and Kumar, Vijay},
  booktitle={52nd IEEE Conference on Decision and Control}, 
  title={Geometric control and differential flatness of a quadrotor UAV with a cable-suspended load}, 
  year={2013},
}

@INPROCEEDINGS{lee2013geometric,
  author={Lee, Taeyoung and Sreenath, Koushil and Kumar, Vijay},
  booktitle={52nd IEEE Conference on Decision and Control}, 
  title={Geometric control of cooperating multiple quadrotor UAVs with a suspended payload}, 
  year={2013},
}

@ARTICLE{lee2018geometric,
  author={Lee, Taeyoung},
  journal={IEEE Transactions on Control Systems Technology}, 
  title={Geometric Control of Quadrotor UAVs Transporting a Cable-Suspended Rigid Body}, 
  year={2018},
}

@ARTICLE{zeng2020differential,
  author={Zeng, Jun and Kotaru, Prasanth and Mueller, Mark W. and Sreenath, Koushil},
  journal={IEEE Robotics and Automation Letters}, 
  title={Differential Flatness Based Path Planning With Direct Collocation on Hybrid Modes for a Quadrotor With a Cable-Suspended Payload}, 
  year={2020},
}

@INPROCEEDINGS{kotaru2017elastic,
  author={Kotaru, Prasanth and Wu, Guofan and Sreenath, Koushil},
  booktitle={2017 American Control Conference (ACC)}, 
  title={Dynamics and control of a quadrotor with a payload suspended through an elastic cable}, 
  year={2017},
}

@INPROCEEDINGS{zeng2019pulley,
  author={Zeng, Jun and Kotaru, Prasanth and Sreenath, Koushil},
  booktitle={2019 American Control Conference (ACC)}, 
  title={Geometric Control and Differential Flatness of a Quadrotor UAV with Load Suspended from a Pulley}, 
  year={2019},
}

@INPROCEEDINGS{kotaru2018flexible,
  author={Kotaru, Prasanth and Wu, Guofan and Sreenath, Koushil},
  booktitle={2018 Indian Control Conference (ICC)}, 
  title={Differential-flatness and control of quadrotor(s) with a payload suspended through flexible cable(s)}, 
  year={2018},
}

@INPROCEEDINGS{tang2015mixed,
  author={Tang, Sarah and Kumar, Vijay},
  booktitle={2015 IEEE International Conference on Robotics and Automation (ICRA)}, 
  title={Mixed Integer Quadratic Program trajectory generation for a quadrotor with a cable-suspended payload},
}

@InProceedings{tang2020multi,
    author="Tang, Sarah and Sreenath, Koushil and Kumar, Vijay",
    editor="Amato, Nancy M. and Hager, Greg and Thomas, Shawna and Torres-Torriti, Miguel",
    title="Multi-robot Trajectory Generation for an Aerial Payload Transport System",
    booktitle="Robotics Research",
    year="2020",
    publisher="Springer International Publishing",
    address="Cham",
    pages="1055--1071",
}

@article{foehn2017fast,
  title={Fast trajectory optimization for agile quadrotor maneuvers with a cable-suspended payload},
  author={Foehn, Philipp and Falanga, Davide and Kuppuswamy, Naveen and Tedrake, Russ and Scaramuzza, Davide},
  year={2017},
  publisher={Robotics: Science and Systems Foundation}
}

@ARTICLE{hwangbo2017control,
  author={Hwangbo, Jemin and Sa, Inkyu and Siegwart, Roland and Hutter, Marco},
  journal={IEEE Robotics and Automation Letters}, 
  title={Control of a Quadrotor With Reinforcement Learning}, 
  year={2017},
}

@article{kaufmann2023champion,
  title={Champion-level drone racing using deep reinforcement learning},
  author={Kaufmann, Elia and Bauersfeld, Leonard and Loquercio, Antonio and M{\"u}ller, Matthias and Koltun, Vladlen and Scaramuzza, Davide},
  journal={Nature},
  year={2023},
}

@article{huang2023datt,
  title={Datt: Deep adaptive trajectory tracking for quadrotor control},
  author={Huang, Kevin and Rana, Rwik and Spitzer, Alexander and Shi, Guanya and Boots, Byron},
  journal={arXiv preprint arXiv:2310.09053},
  year={2023}
}

@ARTICLE{eschmann2024learning,
  author={Eschmann, Jonas and Albani, Dario and Loianno, Giuseppe},
  journal={IEEE Robotics and Automation Letters}, 
  title={Learning to Fly in Seconds}, 
  year={2024}
}

@INPROCEEDINGS{zhang2023learning,
  author={Zhang, Dingqi and Loquercio, Antonio and Wu, Xiangyu and Kumar, Ashish and Malik, Jitendra and Mueller, Mark W.},
  booktitle={2023 IEEE International Conference on Robotics and Automation (ICRA)}, 
  title={Learning a Single Near-hover Position Controller for Vastly Different Quadcopters}, 
  year={2023},
}

@article{zhang2024proxfly,
  title={ProxFly: Robust Control for Close Proximity Quadcopter Flight via Residual Reinforcement Learning},
  author={Zhang, Ruiqi and Zhang, Dingqi and Mueller, Mark W},
  journal={arXiv preprint arXiv:2409.13193},
  year={2024}
}

@article{kaufmann2020deep,
  title={Deep drone acrobatics},
  author={Kaufmann, Elia and Loquercio, Antonio and Ranftl, Ren{\'e} and M{\"u}ller, Matthias and Koltun, Vladlen and Scaramuzza, Davide},
  journal={arXiv preprint arXiv:2006.05768},
  year={2020}
}

@ARTICLE{belkhale2021model,
  author={Belkhale, Suneel and Li, Rachel and Kahn, Gregory and McAllister, Rowan and Calandra, Roberto and Levine, Sergey},
  journal={IEEE Robotics and Automation Letters}, 
  title={Model-Based Meta-Reinforcement Learning for Flight With Suspended Payloads}, 
  year={2021},
}

@ARTICLE{xu2024omnidrones,
  author={Xu, Botian and Gao, Feng and Yu, Chao and Zhang, Ruize and Wu, Yi and Wang, Yu},
  journal={IEEE Robotics and Automation Letters}, 
  title={OmniDrones: An Efficient and Flexible Platform for Reinforcement Learning in Drone Control}, 
  year={2024},
}

@inproceedings{kaufmann2022benchmark,
  title={A benchmark comparison of learned control policies for agile quadrotor flight},
  author={Kaufmann, Elia and Bauersfeld, Leonard and Scaramuzza, Davide},
  booktitle={2022 International Conference on Robotics and Automation (ICRA)},
  pages={10504--10510},
  year={2022},
  organization={IEEE}
}

@article{schulman2017proximal,
  title={Proximal policy optimization algorithms},
  author={Schulman, John and Wolski, Filip and Dhariwal, Prafulla and Radford, Alec and Klimov, Oleg},
}

@article{li2024reinforcement,
author = {Zhongyu Li and Xue Bin Peng and Pieter Abbeel and Sergey Levine and Glen Berseth and Koushil Sreenath},
title ={Reinforcement learning for versatile, dynamic, and robust bipedal locomotion control},
journal = {The International Journal of Robotics Research},
}

@inproceedings{peng2018sim,
  title={Sim-to-real transfer of robotic control with dynamics randomization},
  author={Peng, Xue Bin and Andrychowicz, Marcin and Zaremba, Wojciech and Abbeel, Pieter},
  booktitle={2018 IEEE international conference on robotics and automation (ICRA)},
  pages={3803--3810},
  year={2018},
  organization={IEEE}
}

@article{meuleau2013learning,
  title={Learning finite-state controllers for partially observable environments},
  author={Meuleau, Nicolas and Peshkin, Leonid and Kim, Kee-Eung and Kaelbling, Leslie Pack},
  journal={arXiv preprint arXiv:1301.6721},
  year={2013}
}

@INPROCEEDINGS{tobin2017domain,
  author={Tobin, Josh and Fong, Rachel and Ray, Alex and Schneider, Jonas and Zaremba, Wojciech and Abbeel, Pieter},
  booktitle={2017 IEEE/RSJ International Conference on Intelligent Robots and Systems (IROS)}, 
  title={Domain randomization for transferring deep neural networks from simulation to the real world}, 
  year={2017},
}

@article{cai2024learning,
  title={Learning-based Trajectory Tracking for Bird-inspired Flapping-Wing Robots},
  author={Cai, Jiaze and Sangli, Vishnu and Kim, Mintae and Sreenath, Koushil},
  journal={arXiv preprint arXiv:2411.15130},
  year={2024},
  doi={10.48550/arXiv.2411.15130}
}

@inproceedings{thomas2013avian,
  title={Avian-inspired grasping for quadrotor micro UAVs},
  author={Thomas, Justin and Polin, Joe and Sreenath, Koushil and Kumar, Vijay},
  booktitle={International Design Engineering Technical Conferences and Computers and Information in Engineering Conference},
  year={2013},
}

@book{bertsekas2012dynamic,
  title={Dynamic programming and optimal control: Volume I},
  author={Bertsekas, Dimitri},
  volume={4},
  year={2012},
  publisher={Athena scientific}
}

@article{gupta2025estimation,
  title={Estimation of Aerodynamics Forces in Dynamic Morphing Wing Flight},
  author={Gupta, Bibek and Kim, Mintae and Park, Albert and Sihite, Eric and Sreenath, Koushil and Ramezani, Alireza},
  journal={arXiv preprint arXiv:2508.02984},
  year={2025},
  doi={10.48550/arXiv.2508.02984}
}

@article{bauersfeld2024robotics,
  title={Robotics meets Fluid Dynamics: A Characterization of the Induced Airflow below a Quadrotor as a Turbulent Jet},
  author={Bauersfeld, Leonard and Muller, Koen and Ziegler, Dominic and Coletti, Filippo and Scaramuzza, Davide},
  journal={IEEE Robotics and Automation Letters},
  year={2024},
  publisher={IEEE}
}
\end{document}